\documentclass[11pt,a4paper]{article}
\usepackage[dvipsnames]{xcolor}
\usepackage[hyperref]{acl2019}
\usepackage{times}
\usepackage{latexsym}

\usepackage{booktabs}
\usepackage{url}
\usepackage{verbatim}
\usepackage{multirow}

\usepackage[utf8]{inputenc}
\usepackage{amsmath,amssymb,amsfonts,dsfont,xfrac,physics,xspace}
\usepackage{natbib}
\usepackage{stmaryrd}
\usepackage{tikz}
\usetikzlibrary{decorations.pathmorphing}
\usepackage{trimclip}
\usepackage{graphicx}
\usepackage{subcaption}

\aclfinalcopy
\def\aclpaperid{1599} 

\DeclareMathOperator{\Kuma}{Kuma}

\DeclareMathOperator{\HKuma}{HardKuma}
\DeclareMathOperator{\Cat}{Cat}
\DeclareMathOperator{\Bern}{Bern}
\DeclareMathOperator{\emb}{emb}

\DeclareMathOperator{\rnn}{rnn}

\DeclareMathOperator{\birnn}{birnn}
\DeclareMathOperator{\layer}{layer}
\DeclareMathOperator*{\argmax}{\arg\,\max}
\DeclareMathOperator{\softmax}{softmax}
\DeclareMathOperator{\sparsemax}{sparsemax}
\DeclareMathOperator{\softplus}{softplus}

\newcommand{\vect}[1]{\mathbf{#1}}

\newcommand{\Krv}{\ensuremath{K}\xspace}
\newcommand{\krv}{\ensuremath{k}\xspace}
\newcommand{\Trv}{\ensuremath{T}\xspace}
\newcommand{\trv}{\ensuremath{t}\xspace}
\newcommand{\Hrv}{\ensuremath{H}\xspace}
\newcommand{\hrv}{\ensuremath{h}\xspace}

\definecolor{tblue}{RGB}{31,119,180}
\definecolor{torange}{RGB}{255,127,14}
\definecolor{tgreen}{RGB}{44,160,44}

\title{Interpretable Neural Predictions with Differentiable Binary Variables}

\author{Jasmijn Bastings \\
  ILLC\\University of Amsterdam \\
  {\tt bastings@uva.nl} \\\And
  Wilker Aziz\\
  ILLC\\University of Amsterdam\\
  {\tt w.aziz@uva.nl} \\\And
  Ivan Titov\\
  ILLC, University of Amsterdam\\
  ILCC, University of Edinburgh\\
  {\tt ititov@inf.ed.ac.uk} \\  }

\date{}

\begin{document}

\maketitle
\begin{abstract}
The success of neural networks comes hand in hand with a desire for more interpretability. 
We focus on text classifiers and make them more interpretable by having them provide a justification---a \emph{rationale}---for their predictions. 
We approach this problem by jointly training two neural network models: a latent model that selects a rationale (i.e. a short and informative part of the input text), and a classifier that learns from the words in the rationale alone. 
Previous work proposed to assign binary latent masks to input positions and to promote short selections via sparsity-inducing penalties such as $L_0$ regularisation. 
We propose a latent model that mixes discrete and continuous behaviour allowing at the same time for binary selections and gradient-based training without REINFORCE. In our formulation, we can tractably compute the expected value of penalties such as $L_0$, which allows us to directly optimise the model towards a pre-specified text selection rate. 
We show that our approach is competitive with previous work on rationale extraction, and explore further uses in attention mechanisms.
\end{abstract}
\section{\label{eq:intro}Introduction}

Neural networks are bringing incredible performance gains on text classification tasks \citep{howard2018ulmfit,peters2018elmo,devlin2019bert}. 
However, this power comes hand in hand with a desire for more interpretability, even though its definition may differ \citep{lipton2016mythos}. 
While it is useful to obtain high classification accuracy, with more data available than ever before it also becomes increasingly important to \emph{justify} predictions. Imagine having to classify a large collection of documents, while verifying that the classifications make sense. It would be extremely time-consuming to read each document to evaluate the results. Moreover, if we do not know why a prediction was made, we do not know if we can trust it.\looseness=-1  

What if the model could provide us the most important parts of the document, as a justification for its prediction? 
That is exactly the focus of this paper. We use a setting that was pioneered by \citet{leietal2016rationalizing}.
A \emph{rationale} is defined to be a \emph{short} yet \emph{sufficient} part of the input text; short so that it makes clear what is most important, and sufficient so that a correct prediction can be made from the rationale alone.
One neural network learns to extract the rationale, while another neural network, with separate parameters, learns to make a prediction from just the rationale. 
\citeauthor{leietal2016rationalizing} model this by assigning a binary Bernoulli variable to each input word. The rationale then consists of all the words for which a 1 was sampled. Because gradients do not flow through discrete samples, the rationale extractor is optimized using REINFORCE \citep{Williams1992}. 
An $L_0$ regularizer is used to make sure the rationale is short.

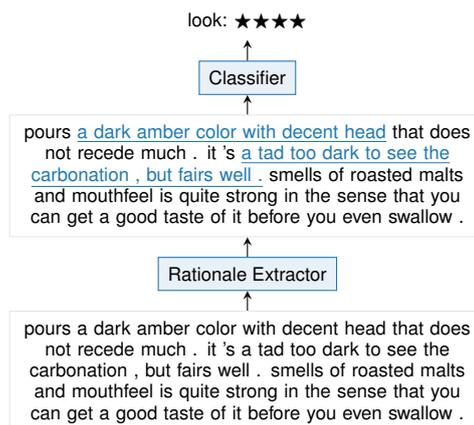
\begin{figure}[t!]
\begin{center}
\begin{tikzpicture}[align=center,font=\sffamily\scriptsize,node distance=1.3cm]
\node [draw=black!10,fill=white,text width=6cm] (input) at (0,0) {pours a dark amber color with decent head that does\\ not recede much . it 's a tad too dark to see the carbonation , but fairs well . smells of roasted malts and mouthfeel is quite strong in the sense that you can get a good taste of it before you even swallow .};
\node [above of=input,draw=tblue,fill=tblue!10] (generator) {Rationale Extractor};
\node [draw=black!10,above of=generator,text width=6cm] (rationale) {pours \textcolor{tblue}{\underline{a dark amber color with decent head}} that does not recede much . it 's \textcolor{tblue}{\underline{a tad too dark to see the} \underline{carbonation , but fairs well .}} smells of roasted malts and mouthfeel is quite strong in the sense that you can get a good taste of it before you even swallow .};
\node [above of=rationale,draw=tblue,fill=tblue!10] (classifier) {Classifier};
\node [above of=classifier,node distance=0.75cm] (result) {look: $\bigstar \bigstar \bigstar \bigstar$};
\path[-stealth] (input) edge (generator);
\path[-stealth] (generator) edge (rationale);
\path[-stealth] (rationale) edge (classifier);
\path[-stealth] (classifier) edge (result);
\end{tikzpicture}
\end{center}
\vspace{-1mm}
    \caption{Rationale extraction for a beer review.}
    \label{fig:latent-rationale-approach}
\end{figure}

We propose an alternative to purely discrete selectors for which gradient estimation is possible without REINFORCE, instead relying on a reparameterization of a random variable that exhibits both continuous and discrete behavior \citep{louizos2017learning}. 
To promote compact rationales, we employ a relaxed form of $L_0$ regularization \citep{louizos2017learning}, penalizing the objective as a function of the expected proportion of selected text. 
We also propose the use of Lagrangian relaxation to target a specific rate of selected input text.

Our contributions are summarized as follows:\footnote{Code available at \url{https://github.com/bastings/interpretable_predictions}.}
\begin{enumerate}
    \item we present a differentiable approach to extractive rationales (\S\ref{sec:latentrationale}) including an objective that allows for specifying how much text is to be extracted (\S\ref{sec:sparsity});
    \item we introduce HardKuma (\S\ref{sec:HKuma}), which gives support to binary outcomes and allows for reparameterized gradient estimates;
    \item we empirically show that our approach is competitive with previous work 
    and that HardKuma has further applications, e.g. in attention mechanisms.
    (\S\ref{sec:experiments}).
\end{enumerate}

\section{Latent Rationale}
\label{sec:latentrationale}
We are interested in making NN-based text classifiers interpretable by (i) uncovering which parts of the input text contribute features for classification, and (ii) basing decisions on only a fraction of the input text (a \emph{rationale}).
\citet{leietal2016rationalizing} approached (i) by inducing binary latent selectors that control which input positions are available to an NN encoder that learns features for classification/regression, and (ii) by regularising their architectures using sparsity-inducing penalties on latent assignments. 
In this section we put their approach under a probabilistic light, and this will then more naturally lead to our proposed method. 

In text classification, an input $x$ is mapped to a distribution over target labels: 
\begin{equation}
    Y|x \sim \Cat(f(x; \theta)) ~,
\end{equation}
where we have a neural network architecture $f(\cdot; \theta)$ parameterize the model---$\theta$ collectively denotes the parameters of the NN layers in $f$. That is, an NN maps from data space (e.g. sentences, short paragraphs, or premise-hypothesis pairs) to the categorical parameter space (i.e. a vector of class probabilities). 
For the sake of concreteness, consider the input a sequence $x = \langle x_1, \ldots, x_n\rangle$.  
A target $y$ is typically a categorical outcome, such as a sentiment class or an entailment decision, but with an appropriate choice of likelihood it could also be a numerical score (continuous or integer).

\citet{leietal2016rationalizing} augment this model with a collection of latent variables  which we denote by $z = \langle z_1, \ldots, z_n \rangle$. These variables are responsible for regulating which portions of the input $x$ contribute with predictors (i.e. features) to the classifier. 
The model formulation changes as follows:
\begin{equation}
\begin{aligned}
    Z_i|x &\sim \Bern(g_i(x; \phi)) \\
    Y|x, z &\sim \Cat(f(x \odot z; \theta)) \label{eq:gated_obs_model}
\end{aligned}
\end{equation}
where an NN $g(\cdot; \phi)$ predicts a sequence of $n$ Bernoulli parameters---one per latent variable---and the classifier is modified such that $z_i$ indicates whether or not $x_i$ is available for encoding. We can think of the sequence $z$ as a  binary gating mechanism used to select a rationale, which with some abuse of notation we denote by $x \odot z$. 
Figure \ref{fig:latent-rationale-approach} illustrates the approach.

Parameter estimation for this model can be done by maximizing a lower bound $\mathcal E(\phi, \theta)$ on the log-likelihood of the data derived by application of Jensen's inequality:\footnote{This can be seen as variational inference \citep{Jordan+1999:VI} where we perform approximate inference using a data-dependent prior $P(z|x, \phi)$.}
\begin{equation}
\begin{aligned}\label{eq:lowerbound}
    &\log P(y|x) = \log \mathbb E_{P(z|x, \phi)}\left[ P(y|x, z, \theta) \right] \\ 
    &\overset{\text{JI}}{\ge} \mathbb E_{P(z|x, \phi)}\left [\log P(y|x, z, \theta) \right] = \mathcal E(\phi, \theta)  ~ .
\end{aligned}
\end{equation}

These latent rationales approach the first objective, namely, uncovering which parts of the input text contribute towards a decision. 
However note that an NN controls the Bernoulli parameters, thus nothing prevents this NN from selecting the whole of the input, thus defaulting to a standard text classifier.
To promote compact rationales, \citet{leietal2016rationalizing} impose sparsity-inducing penalties on latent selectors. 
They penalise for the total number of selected words, $L_0$ in (\ref{eq:lei-objective}), as well as, for the total number of transitions, \emph{fused lasso} in (\ref{eq:lei-objective}), and approach the following optimization problem 
\begin{equation}\label{eq:lei-objective}
    \min_{\phi, \theta} - \mathcal E(\phi, \theta) + \lambda_0 \! \underbrace{\sum_{i=1}^n z_i}_{L_0(z)} + \lambda_1 \! \underbrace{\sum_{i=1}^{n-1} |z_i - z_{i+1}|}_{\text{fused lasso}}
\end{equation}
via gradient-based optimisation, where $\lambda_0$ and $\lambda_1$ are fixed hyperparameters.
The objective is however intractable to compute, the lowerbound, in particular, requires marginalization of $O(2^n)$ binary sequences.
For that reason, \citeauthor{leietal2016rationalizing} sample latent assignments and work with  gradient estimates using REINFORCE \citep{Williams1992}.

The key ingredients are, therefore, binary latent variables and sparsity-inducing regularization, and therefore the solution is marked by non-differentiability. 
We propose to replace Bernoulli variables by rectified continuous random variables \citep{RectGaussian}, for they exhibit both discrete and continuous behaviour.
Moreover, they are amenable to reparameterization in terms of a fixed random source \citep{Kingma+2014:VAE}, in which case gradient estimation is possible without REINFORCE.
Following \citet{louizos2017learning}, we exploit one such distribution to relax $L_0$ regularization and thus promote compact rationales with a differentiable objective.
In section \ref{sec:HKuma}, we introduce this distribution and present its properties. In section \ref{sec:sparsity}, we employ a Lagrangian relaxation to automatically target a pre-specified selection rate.
And finally, in section \ref{sec:model} we present an example for sentiment classification.

\section{\label{sec:HKuma}Hard Kumaraswamy Distribution}

Key to our model is a novel distribution that exhibits both continuous and discrete behaviour, in this section we introduce it.
With non-negligible probability, samples from this distribution evaluate to exactly $0$ or exactly $1$.
In a nutshell: i) we start from a distribution over the open interval $(0, 1)$ (see dashed curve in Figure \ref{fig:kuma}); ii) we then \emph{stretch} its support from $l < 0$ to $r > 1$ in order to include $\{0\}$ and $\{1\}$ (see solid curve in Figure \ref{fig:kuma}); finally,  iii) we collapse the probability mass over the interval $(l, 0]$ to $\{0\}$, and similarly, the probability mass over the interval $[1, r)$ to $\{1\}$ (shaded areas in Figure \ref{fig:kuma}).
This \emph{stretch-and-rectify} technique was proposed by \citet{louizos2017learning}, who rectified samples from the BinaryConcrete (or GumbelSoftmax) distribution \citep{MaddisonEtAl2017:Concrete,JangEtAl2017:GumbelSoftmax}.
We adapted their technique to the Kumaraswamy distribution motivated by its close resemblance to a Beta distribution, for which we have stronger intuitions (for example, its two shape parameters transit rather naturally from unimodal to bimodal configurations of the distribution).
In the following, we introduce this new distribution formally.\footnote{We use uppercase letters for random variables (e.g. \Krv, \Trv, and \Hrv) and lowercase for assignments (e.g. \krv, \trv, \hrv). For a random variable $\Krv$, $f_\Krv(\krv; \alpha)$ is the probability density function (pdf), conditioned on parameters $\alpha$, and $F_\Krv(\krv; \alpha)$ is the cumulative distribution function (cdf).} 

\subsection{Kumaraswamy distribution} 

\begin{figure}[tb]
    \begin{center}
        \includegraphics[width=\columnwidth]{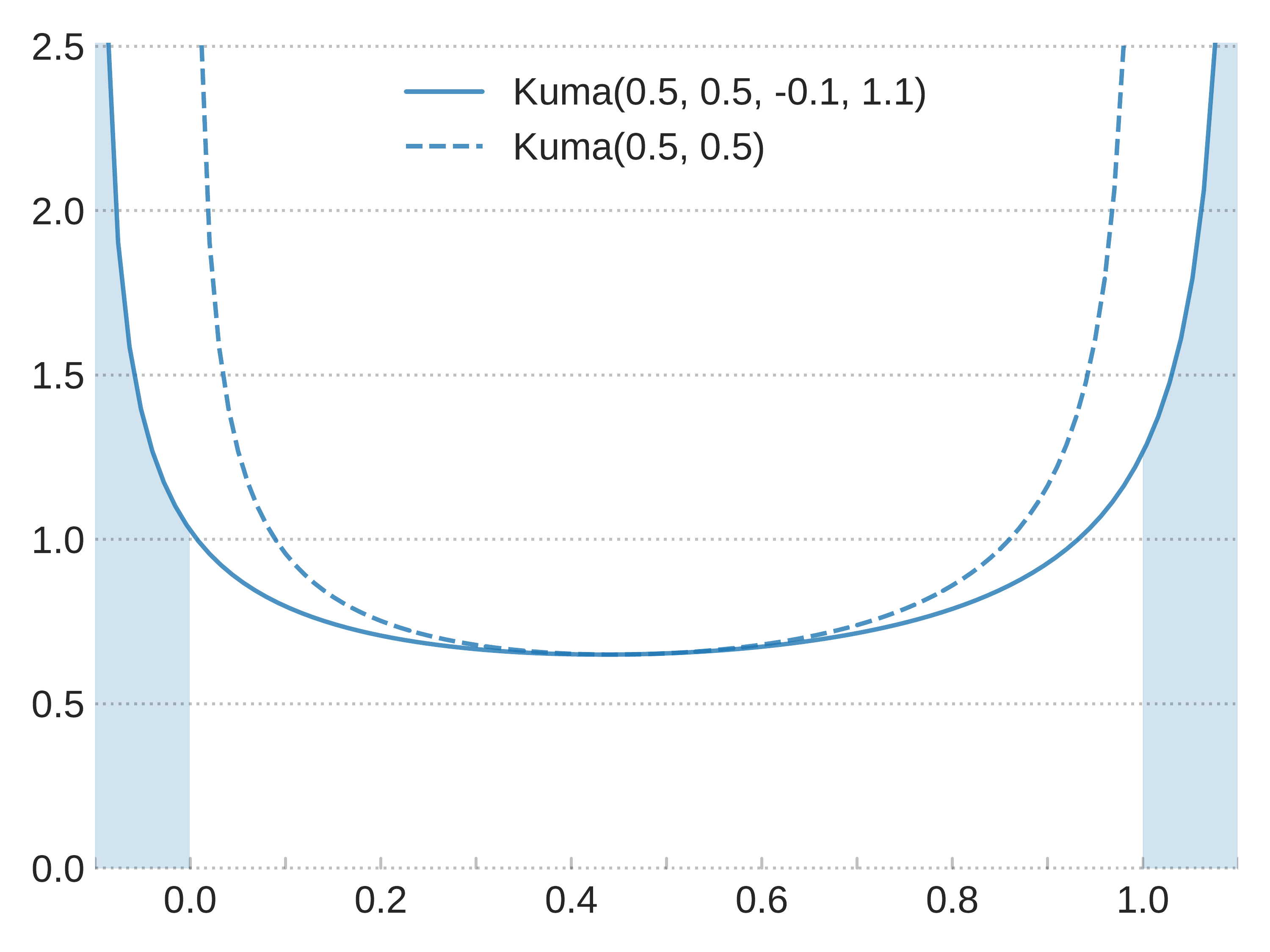}
    \end{center}
    \vspace{-5mm}
    \caption{The $\HKuma$ distribution: we start from a $\Kuma(0.5, 0.5)$, and stretch its support to the interval $(-0.1, 1.1)$, finally we collapse all mass before $0$ to $\{0\}$ and all mass after $1$ to $\{1\}$.}
    \label{fig:kuma}
\end{figure}

The Kumaraswamy distribution \citep{kumaraswamy1980generalized} is a two-parameters distribution over the {\bf open} interval $(0, 1)$, we denote a Kumaraswamy-distributed variable by $\Krv \sim \Kuma(a, b)$, where $a \in \mathbb R_{>0}$ and $b \in \mathbb R_{>0}$ control the distribution's shape.  The dashed curve in Figure \ref{fig:kuma} illustrates the density of $\Kuma(0.5, 0.5)$. 
For more details including its pdf and cdf, consult Appendix \ref{app:kuma}.

The Kumaraswamy is a close relative of the Beta distribution, though not itself an exponential family, with a simple cdf whose inverse
\begin{equation}
    F_\Krv^{-1}(u; a, b) = \left(1- (1-u)^{\sfrac{1}{b}}\right)^{\sfrac{1}{a}} ~,
\end{equation}
for $u \in [0, 1]$, can be used to obtain samples
\begin{equation}
    F_K^{-1}(U; \alpha, \beta) \sim \Kuma(\alpha, \beta) 
\end{equation}
by transformation of a uniform random source $U \sim \mathcal U(0, 1)$.
We can use this fact to reparameterize expectations \citep{nalisnick2016stick}.  

\subsection{Rectified Kumaraswamy} 
We \emph{stretch} the support of the Kumaraswamy distribution to include $0$ and $1$. The resulting variable $\Trv \sim \Kuma(a, b, l, r)$ takes on values in the open interval $(l, r)$ where $l < 0$ and $r > 1$, with cdf 
\begin{equation}
F_\Trv(\trv; a, b, l, r) = F_\Krv(\sfrac{(\trv - l)}{(r-l)}; a, b)   ~ . \label{eq:trv_cdf}    
\end{equation}

We now define a rectified random variable, denoted by $\Hrv \sim \HKuma(a, b, l, r)$, by passing a sample $\Trv \sim \Kuma(a, b, l, r)$ through a hard-sigmoid, i.e. $\hrv =  \min(1, \max(0, \trv))$. 
The resulting variable is defined over the {\bf closed} interval $[0, 1]$.
Note that while there is $0$ probability of sampling $\trv = 0$, 
sampling $\hrv = 0$ corresponds to sampling any $\trv \in (l, 0]$, a set whose mass under $\Kuma(\trv|a, b, l, r)$ is available in closed form: 
\begin{equation}
    \mathbb P(\Hrv = 0) = F_\Krv\left(\textstyle\frac{-l}{r-l}; a, b\right) ~.
\end{equation}
That is because all negative values of $\trv$ are deterministically mapped to zero.
Similarly, samples $\trv \in [1, r)$ are all deterministically mapped to $\hrv = 1$, whose total mass amounts to 
\begin{equation}
    \mathbb P(\Hrv=1) = 1 - F_\Krv\left(\textstyle\frac{1-l}{r-l}; a, b\right) ~.
\end{equation}
See Figure \ref{fig:kuma} for an illustration, and Appendix \ref{app:kuma} for the complete derivations.

\subsection{Reparameterization and gradients}

Because this rectified variable is built upon a Kumaraswamy, it admits a reparameterisation in terms of a uniform variable $U \sim \mathcal U(0, 1)$. We need to first sample a uniform variable in the open interval $(0, 1)$ and transform the result to a Kumaraswamy variable via the inverse cdf (\ref{eq:u-to-k}), then shift and scale the result to cover the stretched support (\ref{eq:k-to-t}), and finally, apply the rectifier in order to get a sample in the closed interval $[0, 1]$ (\ref{eq:t-to-h}).
\begin{subequations}
\begin{align}
    \krv &= F^{-1}_\Krv(u; a, b) \label{eq:u-to-k}\\
    \trv &= l + (r-l)\krv  \label{eq:k-to-t}\\
    \hrv &= \min(1, \max(0, \trv)) ~, \label{eq:t-to-h} 
\end{align}
\end{subequations}
We denote this $\hrv = s(u; a, b, l, r)$ for short.
Note that this transformation has two discontinuity points, namely, $\trv = 0$ and $\trv = 1$. 
Though recall, the probability of sampling $\trv$ exactly $0$ or exactly $1$ is zero, which essentially means stochasticity circumvents points of non-differentiability of the rectifier (see Appendix \ref{app:grad}).


\section{Controlled Sparsity}
\label{sec:sparsity}

Following \citet{louizos2017learning}, we relax non-differentiable penalties by computing them on expectation under our latent model $p(z|x, \phi)$. 
In addition, we propose the use of Lagrangian relaxation to target specific values for the penalties. 
Thanks to the tractable Kumaraswamy cdf, the expected value of $L_0(z)$ is known in closed form
\begin{equation}
\begin{aligned}
    &\mathbb E_{p(z|x)}\left[ L_0(z) \right] \overset{\text{ind}}{=}  \sum_{i=1}^n \mathbb E_{p(z_i|x)}\left[ \mathbb I[z_i \neq 0] \right] \\
    &=  \sum_{i=1}^n 1 - \mathbb P(Z_i = 0) ~,
\end{aligned}
\end{equation}
where $\mathbb P(Z_i=0) = F_K\left(\frac{-l}{r-l}; a_i, b_i\right)$.
This quantity is a tractable and differentiable function of the parameters $\phi$ of the latent model.
We can also compute a relaxation of fused lasso by computing the expected number of zero-to-nonzero and nonzero-to-zero changes:
\begin{equation}
\begin{aligned}
    &\mathbb E_{p(z|x)}\left[ \sum_{i=1}^{n-1}  \mathbb I[z_i = 0, z_{i+1}\neq 0] \right] \\
    &+ \mathbb E_{p(z|x)}\left[ \sum_{i=1}^{n-1}  \mathbb I[z_i \neq 0, z_{i+1}= 0] \right] \\
   &\overset{\text{ind}}{=}  \sum_{i=1}^{n-1} \mathbb P(Z_i = 0)(1 - \mathbb P(Z_{i+1} = 0)) \\
   &+ (1 - \mathbb P(Z_i = 0))\mathbb P(Z_{i+1} = 0) ~ .
\end{aligned}
\end{equation}
In both cases, we make the assumption that latent variables are independent given $x$, in Appendix \ref{app:depz} we discuss how to estimate the regularizers for a model $p(z_i|x, z_{<i})$ that conditions on the prefix $z_{<i}$ of sampled $\HKuma$ assignments.



We can use regularizers to promote sparsity, but just how much text will our final model select?
Ideally, we would target specific values $r$ and solve a constrained optimization problem. 
In practice, constrained optimisation is very challenging, thus  
we employ Lagrangian relaxation instead: 
\begin{equation}\label{eq:our-objective}
\max_{\lambda \in \mathbb R} \min_{\phi,\theta} -\mathcal{E}(\phi, \theta) + \lambda^\top (R(\phi) - r)
\end{equation}
\noindent where $R(\phi)$ is a vector of regularisers, e.g. expected $L_0$ and expected fused lasso, and $\lambda$ is a vector of Lagrangian multipliers $\lambda$.
Note how this differs from the treatment of \citet{leietal2016rationalizing} shown in (\ref{eq:lei-objective}) where regularizers are computed for assignments, rather than on expectation, and where $\lambda_0, \lambda_1$ are fixed hyperparameters.

\section{\label{sec:model}Sentiment Classification}

As a concrete example, consider the case of sentiment classification where $x$ is a sentence and $y$ is a $5$-way sentiment class (from very negative to very positive).  
The model consists of 
\begin{equation}
\begin{aligned}
    Z_i &\sim \HKuma(a_i, b_i, l, r) \\
    Y|x,z &\sim \Cat(f(x \odot z; \theta))
\end{aligned}
\end{equation}
where the shape parameters $a, b = g(x; \phi)$, i.e. two sequences of $n$ strictly positive scalars, are predicted by a NN, and the support boundaries $(l, r)$ are fixed hyperparameters.

We first specify an architecture that parameterizes latent selectors and then use a reparameterized sample to restrict which parts of the input contribute encodings for classification:\footnote{We describe architectures using blocks denoted by  $\layer(\text{inputs}; \text{subset of parameters})$, boldface letters for vectors, and the shorthand $\vect{v}_1^n$ for a sequence $\langle \vect{v}_1, \ldots, \vect{v}_n\rangle$.}
\begin{equation*}
\begin{split}
    \mathbf e_i &= \emb(x_i) \\
    \vect{h}_1^n &= \birnn(\vect{e}_1^n; \phi_r)\\  
    u_i &\sim \mathcal U(0, 1) \\
\end{split}
\quad
\begin{split}
    a_i &= f_a(\vect{h}_i; \phi_a) \\
    b_i &= f_b(\vect{h}_i; \phi_b) \\
    z_i &= s(u_i; a_i, b_i, l, r)
\end{split}
\end{equation*}
where $\emb(\cdot)$ is an embedding layer, $\birnn(\cdot;\phi_r)$ is a bidirectional encoder, $f_a(\cdot; \phi_a)$ and $f_b(\cdot; \phi_b)$ are feed-forward transformations with $\softplus$ outputs, and $s(\cdot)$ turns the uniform sample $u_i$ into the latent selector $z_i$ (see \S\ref{sec:HKuma}).
We then use the sampled $z$ to modulate inputs to the classifier:  
\begin{equation*}
\begin{aligned}
    \vect{e}_i &= \emb(x_i)\\
    \vect{h}_i^{(\text{fwd})} &= \rnn(\vect{h}_{i-1}^{(\text{fwd})}, z_i \, \vect{e}_i; \theta_{\text{fwd}}) \\ 
    \vect{h}_i^{(\text{bwd})} &= \rnn(\vect{h}_{i+1}^{(\text{bwd})},  z_i \, \vect{e}_i; \theta_{\text{bwd}}) \\ 
    \vect{o} &= f_o(\vect{h}_n^{(\text{fwd})}, \vect{h}_1^{(\text{bwd})}; \theta_o) 
\end{aligned}
\end{equation*}
where $\rnn(\cdot; \theta_{\text{fwd}})$ and  $\rnn(\cdot; \theta_{\text{bwd}})$ are recurrent cells such as LSTMs \citep{hochreiter97} that process the sequence in different directions, and $f_o(\cdot; \theta_o)$ is a feed-forward transformation with $\softmax$ output.
Note how $z_i$ modulates features $\mathbf e_i$ of the input $x_i$ that are available to the recurrent composition function. 

We then obtain gradient estimates of $\mathcal E(\phi, \theta)$ via Monte Carlo (MC) sampling from 
\begin{equation}\label{eq:dtheta}
\begin{aligned}
\mathcal E(\phi, \theta) = \mathbb E_{\mathcal U(0, I)}\left[ \log P(y|x, s_\phi(u, x), \theta)\right] \\
\end{aligned}
\end{equation}
where $z = s_\phi(u, x)$ is a shorthand for elementwise application of the transformation from uniform samples to HardKuma samples. 
This reparameterisation is the key to gradient estimation through stochastic computation graphs \citep{Kingma+2014:VAE,RezendeEtAl14VAE}. 

\paragraph{Deterministic predictions.} At test time we make predictions based on what is the most likely assignment for each $z_i$. We $\argmax$ across configurations of the distribution, namely, $z_i = 0$, $z_i=1$, or $0 < z_i < 1$.
When the continuous interval is more likely, we take the expected value of the underlying Kumaraswamy variable. 
\section{Experiments}
\label{sec:experiments}

We perform experiments on multi-aspect sentiment analysis to compare with previous work, as well as experiments on sentiment classification and natural language inference. All models were implemented in PyTorch, and Appendix~\ref{app:implementationdetails} provides implementation details.

\paragraph{Goal.} When rationalizing predictions, our goal is to perform as well as systems using the full input text, while using only a \textit{subset} of the input text, leaving unnecessary words out for interpretability.

\begin{table}[tb]
    \centering
    \begin{tabular}{lr}
    \toprule
    SVM \citep{leietal2016rationalizing}    & 0.0154 \\
    BiLSTM \citep{leietal2016rationalizing}   & 0.0094 \\
    BiRCNN \citep{leietal2016rationalizing}   & 0.0087 \\
    \midrule
    BiLSTM \textit{(ours)}                           & 0.0089 \\
    BiRCNN \textit{(ours)}                           & 0.0088 \\
    \bottomrule
    \end{tabular}
    \caption{MSE on the BeerAdvocate test set.}
    \label{tab:beer-mse}
\end{table}

\subsection{Multi-aspect Sentiment Analysis}

\begin{table*}[t]
    \centering
    \begin{tabular}{lrrrrrr}
    \toprule
    \multirow{2}[3]{*}{Method}  & \multicolumn{2}{c}{Look} & \multicolumn{2}{c}{Smell} & \multicolumn{2}{c}{Palate} \\
     \cmidrule(lr){2-3} \cmidrule(lr){4-5} \cmidrule(lr){6-7}
            & {\small\% Precision} & {\small \% Selected} & {\small \% Precision} & {\small \% Selected}  & {\small \% Precision} & {\small \% Selected}  \\
    \midrule
    Attention (\citeauthor{leietal2016rationalizing})           & 80.6 & 13      & 88.4 & 7      & 65.3 & 7\\
    Bernoulli (\citeauthor{leietal2016rationalizing})    & 96.3 & 14      & 95.1 & 7      & 80.2 & 7\\
    Bernoulli \emph{(reimpl.)} & 94.8 & 13 & 95.1 & 7 & 80.5 & 7\\
    HardKuma  & 98.1 & 13 & 96.8 & 7 & 89.8 & 7\\
    \bottomrule
    \end{tabular}
    \caption{Precision (\% of selected words that was also annotated as the gold rationale) and selected (\% of words not zeroed out) per aspect.
    In the attention baseline, the top 13\% (7\%) of words with highest attention weights are used for classification. 
    Models were selected based on validation loss.}
    \label{tab:beer-results}
\end{table*}

In our first experiment we compare directly with previous work on rationalizing predictions \citep{leietal2016rationalizing}. We replicate their setting.

\paragraph{Data.} A pre-processed subset of the BeerAdvocate\footnote{\url{https://www.beeradvocate.com/}} data set is used \citep{mcauley2012learning}. It consists of 220,000 \textit{beer reviews}, where multiple aspects (e.g. look, smell, palate) are rated.
As shown in Figure~\ref{fig:latent-rationale-approach}, a review typically consists of multiple sentences, and contains a 0-5 star rating (e.g. 3.5 stars) for each aspect. 
\citeauthor{leietal2016rationalizing} mapped the ratings to scalars in $[0, 1]$.

\paragraph{Model.} We use the models described in \S\ref{sec:model} with two small modifications: 1) since this is a regression task, we use a sigmoid activation in the output layer of the classifier rather than a softmax,\footnote{From a likelihood learning point of view, we would have assumed a Logit-Normal likelihood, however, to stay closer to \citet{leietal2016rationalizing}, we employ mean squared error.} 
and 2) we use an extra RNN to condition $z_i$ on $z_{<i}$:
\begin{subequations}
\begin{align}
    a_i &= f_a(\vect{h}_i, \vect{s}_{i-1}; \phi_a) \\
    b_i &= f_b(\vect{h}_i, \vect{s}_{i-1}; \phi_b) \\
    \vect{s}_i &= \text{rnn}(\vect{h}_i, z_i, \vect{s}_{i-1} ; \phi_s )
\end{align}
\end{subequations}
For a fair comparison we follow \citeauthor{leietal2016rationalizing} by using RCNN\footnote{An RCNN cell can replace any LSTM cell and works well on text classification problems. See appendix~\ref{app:implementationdetails}.} cells rather than LSTM cells for encoding sentences on this task. 
Since this cell is not widely used, we verified its performance in Table~\ref{tab:beer-mse}.  We observe that the BiRCNN performs on par with the BiLSTM (while using 50\% fewer parameters), and similarly to previous results. 

\paragraph{Evaluation.} A test set with sentence-level rationale annotations is available. The \emph{precision} of a rationale is defined as the percentage of words with $z \ne 0$ that is part of the annotation. We also evaluate the predictions made from the rationale using mean squared error (MSE).

\paragraph{Baselines.} For our baseline we reimplemented the approach of \citet{leietal2016rationalizing} which we call \emph{Bernoulli} after the distribution they use to sample $z$ from. We also report their attention baseline, in which an attention score is computed for each word, after which it is simply thresholded to select the top-k percent as the rationale.

\begin{figure}[tb]
    \vspace{-2mm}
    \begin{center}
    \includegraphics[width=\columnwidth]{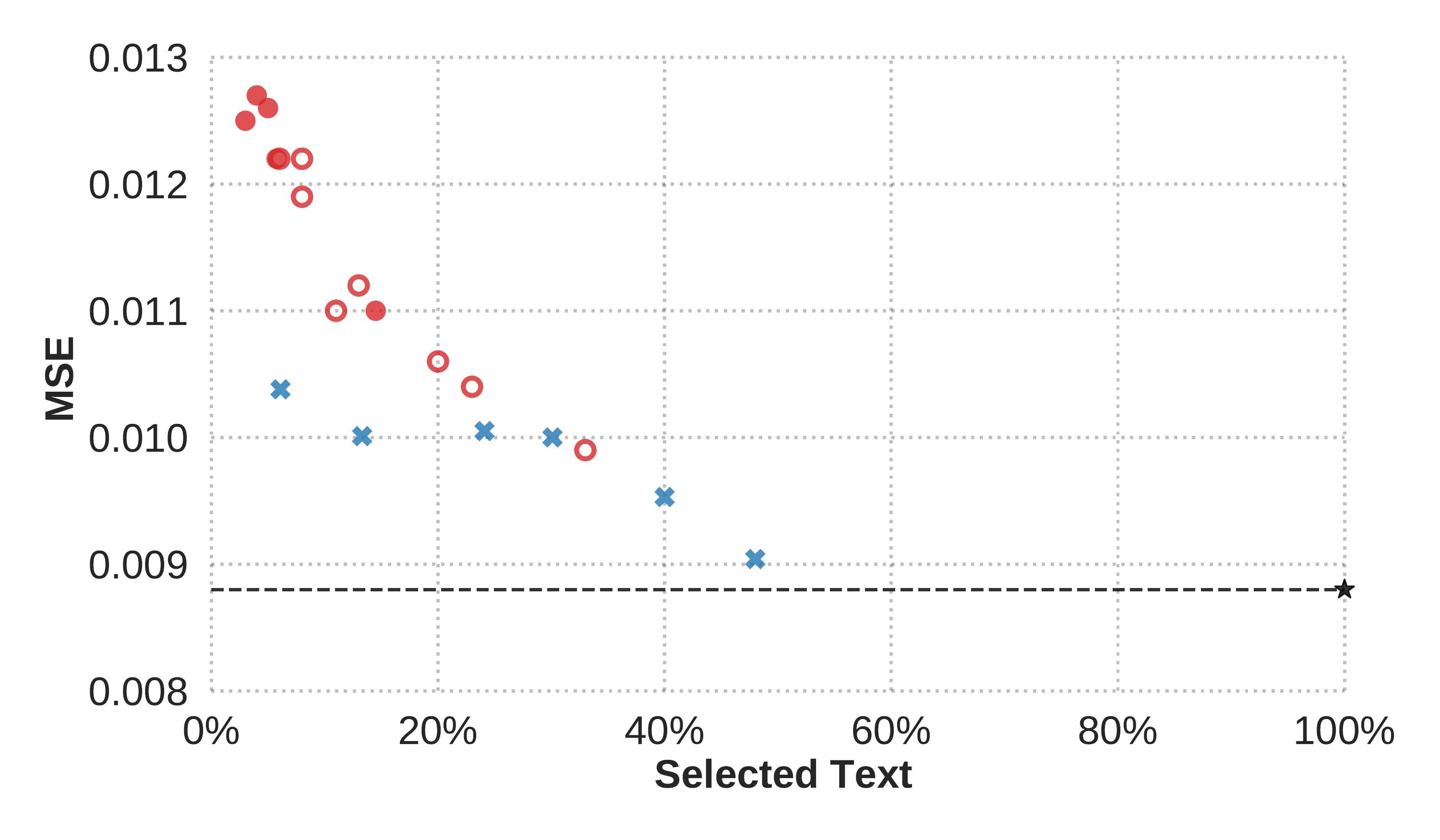}
    \end{center}    
    \vspace{-4mm}
    \caption{MSE of all aspects for various percentages of extracted text. 
    HardKuma (blue crosses) has lower error than Bernoulli (red circles; open circles taken from \citet{leietal2016rationalizing}) for similar amount of extracted text. The full-text baseline (black star) gets the best MSE.}
    \label{fig:beer-tradeoff}
\end{figure}

\paragraph{Results.} Table~\ref{tab:beer-results} shows the precision and the percentage of selected words for the first three aspects. 
The models here have been selected based on validation MSE and were tuned to select a similar percentage of words (`selected'). 
We observe that our Bernoulli reimplementation reaches the precision similar to previous work, doing a little bit worse for the `look' aspect. 
Our HardKuma managed to get even higher precision, and it extracted exactly the percentage of text that we specified (see \S\ref{sec:sparsity}).\footnote{We tried to use Lagrangian relaxation for the Bernoulli model, but this led to instabilities (e.g. all words selected).} 
Figure~\ref{fig:beer-tradeoff} shows the MSE for all aspects for various percentages of extracted text. We observe that HardKuma does better with a smaller percentage of text selected. The performance becomes more similar as more text is selected.

\subsection{Sentiment Classification}

We also experiment on the Stanford Sentiment Treebank (SST) \citep{socheretal2013sst}. 
There are 5 sentiment classes: very negative, negative, neutral, positive, and very positive.
Here we use the HardKuma model described in \S\ref{sec:model}, a Bernoulli model trained with REINFORCE, as well as a BiLSTM.

\begin{figure}[tb]
    \vspace{-0mm}
    \begin{center}
    \includegraphics[width=\columnwidth]{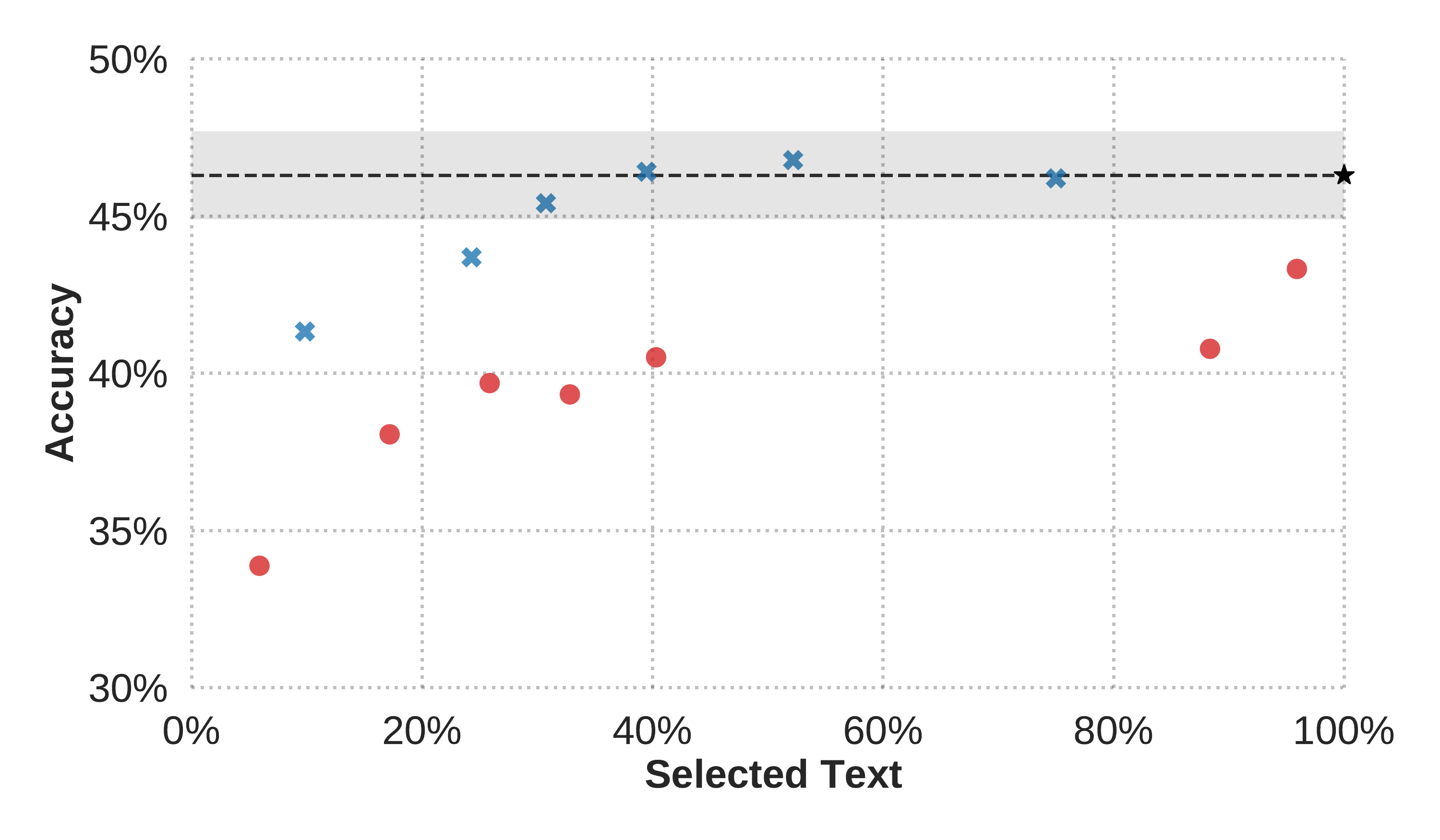}
    \end{center}    
    \vspace{-5mm}
    \caption{SST validation accuracy for various percentages of extracted text. 
    HardKuma (blue crosses) has higher accuracy than Bernoulli (red circles) for similar amount of  text, and reaches the full-text baseline (black star, $46.3 \pm 2\sigma$ with $\sigma=0.7$) around 40\% text.}
    \label{fig:sst-tradeoff}
\end{figure}

\paragraph{Results.} Figure~\ref{fig:sst-tradeoff} shows the classification accuracy for various percentages of selected text. We observe that HardKuma outperforms the Bernoulli model at each percentage of selected text. HardKuma reaches full-text baseline performance already around 40\% extracted text. 
At that point, it obtains a \emph{test} score of 45.84, versus 42.22 for Bernoulli and 47.4$\pm0.8$ for the full-text baseline.

\begin{figure}[tb]
    \vspace{-4mm}
    \begin{center}
    \includegraphics[width=\columnwidth,clip,trim=5mm 1cm 0 5mm]{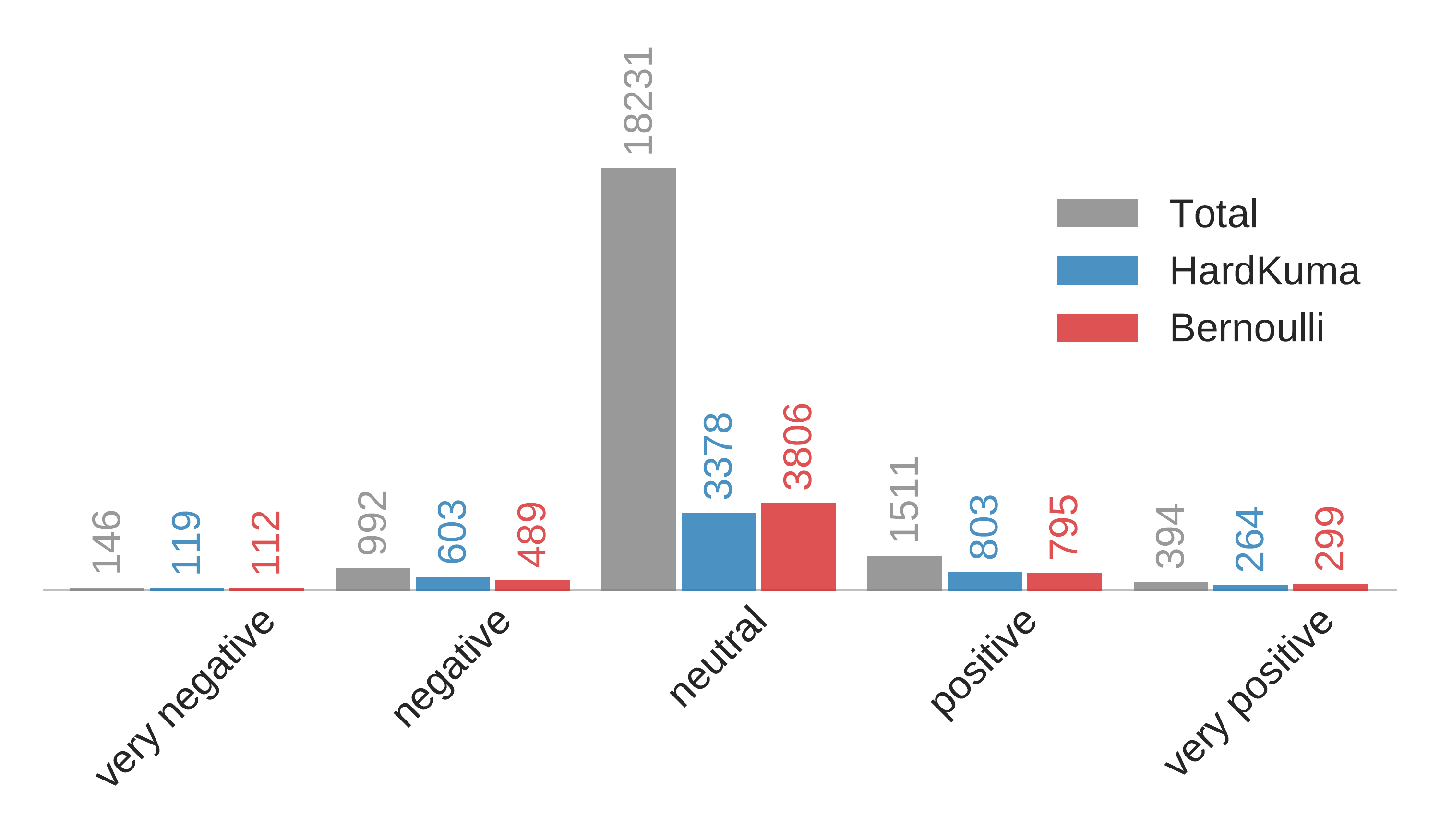}
    \end{center}    
    \vspace{-4mm}
    \caption{The number of words in each sentiment class for the full validation set, the HardKuma (24\% selected text) and Bernoulli (25\% text).}
    \label{fig:sst-hist}
\end{figure}

\paragraph{Analysis.} We wonder what kind of words are dropped when we select smaller amounts of text. For this analysis we exploit the word-level sentiment annotations in SST, which allows us to track the sentiment of words in the rationale. Figure~\ref{fig:sst-hist} shows that a large portion of dropped words have neutral sentiment, and it seems plausible that exactly those words are not important features for classification. We also see that HardKuma drops (relatively) more neutral words than Bernoulli.

\subsection{Natural Language Inference}
In Natural language inference (NLI), given a premise sentence $x^{(p)}$ and a hypothesis sentence $x^{(h)}$, the goal is to predict their relation $y$ which can be contradiction, entailment, or neutral. As our dataset we use the Stanford Natural Language Inference (SNLI) corpus \citep{bowman2015nli}.

\paragraph{Baseline.} We use the Decomposable Attention model (DA) of \citet{parikhetal2016decomposable}.\footnote{Better results e.g. \citet{chenetal2017esim} and data sets for NLI exist, but are not the focus of this paper.}
DA does not make use of LSTMs, but rather uses attention to find connections between the premise and the hypothesis that are predictive of the relation. Each word in the premise attends to each word in the hypothesis, and vice versa, resulting in a set of comparison vectors which are then aggregated for a final prediction. If there is no link between a word pair, it is not considered for prediction.

\paragraph{Model.} 
Because the premise and hypothesis interact, it does not make sense to extract a rationale for the premise and hypothesis independently. 
Instead, we replace the attention between premise and hypothesis with HardKuma attention. Whereas in the baseline a similarity matrix is $\softmax$-normalized across rows (premise to hypothesis) and columns (hypothesis to premise) to produce attention matrices, in our model each cell in the attention matrix is sampled from a HardKuma parameterized by $(a, b)$. 
To promote sparsity, we use the relaxed $L_0$ to specify the desired percentage of non-zero attention cells.
The resulting matrix does not need further normalization.

\paragraph{Results.} With a target rate of 10\%, the HardKuma model achieved 8.5\% non-zero attention. Table~\ref{tab:snli-results} shows that, even with so many zeros in the attention matrices, it only does about 1\% worse compared to the DA baseline. Figure~\ref{fig:snli-hardkuma-example} shows an example of HardKuma attention, with additional examples in Appendix~\ref{app:implementationdetails}. We leave further explorations with HardKuma attention for future work.

\begin{table}[hb]
    \centering
    \begin{tabular}{lcc}
        \toprule 
        Model & Dev & Test \\
        \midrule
        LSTM \citep{bowman2016unified}           & -- & 80.6 \\
        DA \citep{parikhetal2016decomposable}    & -- & 86.3 \\
        \midrule
        DA \emph{(reimplementation)}            & 86.9 & 86.5\\
        DA with HardKuma attention           & 86.0 & 85.5 \\
        \bottomrule
    \end{tabular}
    \caption{SNLI results (accuracy).}
    \label{tab:snli-results}
\end{table}
%
\begin{figure}[b!]
    \vspace{-10mm}
    \begin{center}
    \includegraphics[width=\columnwidth,clip,trim=0 5.2cm 0 2.65cm]{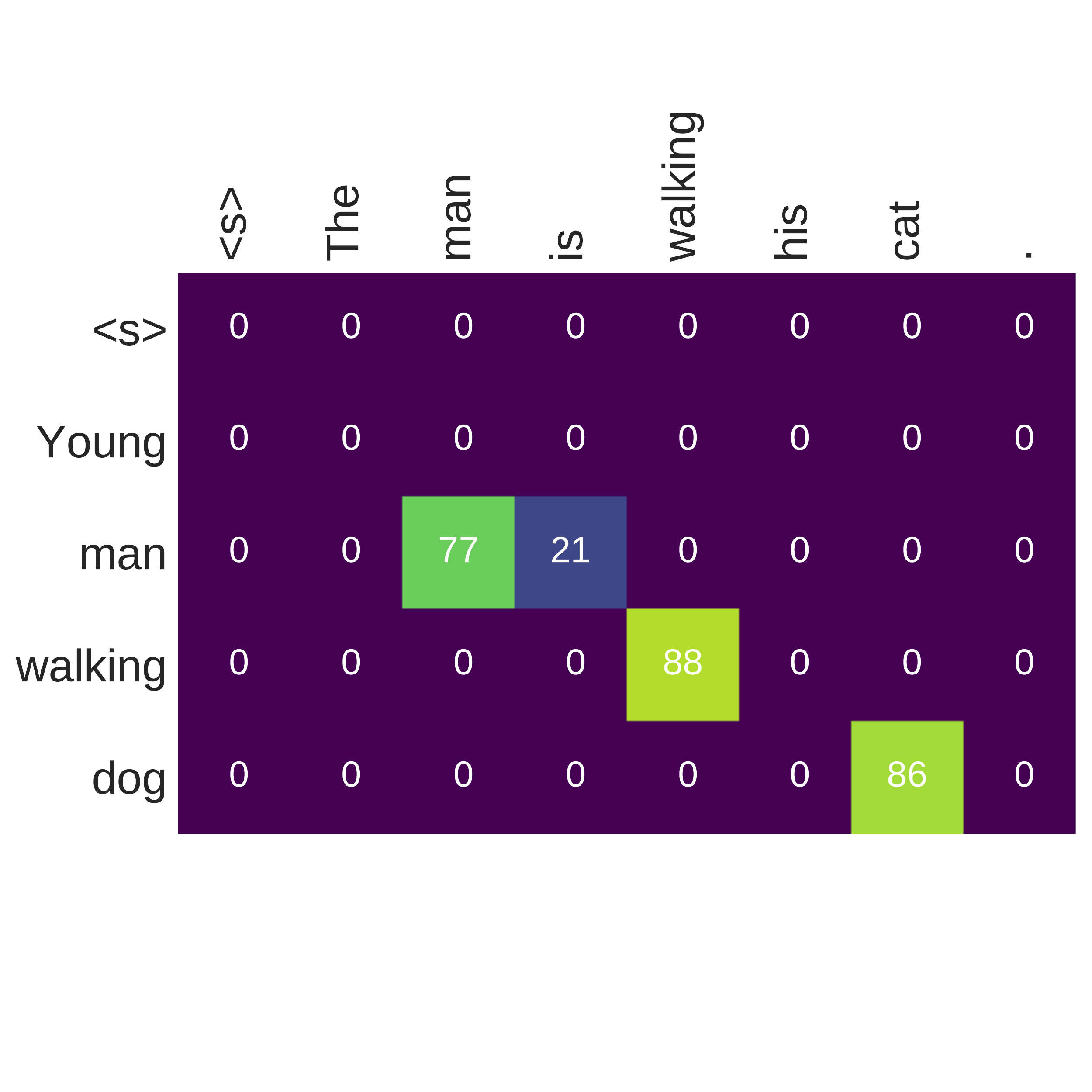}
    \end{center}    
    \vspace{-6mm}
    \caption{Example of HardKuma attention between a premise (rows) and hypothesis (columns) in  SNLI (cell values shown in multiples of $10^{-2}$).}
    \label{fig:snli-hardkuma-example}
\end{figure}

\section{Related Work}
This work has connections with work on interpretability, learning from rationales, sparse structures, and rectified distributions. 
We discuss each of those areas.

\paragraph{Interpretability.} 
Machine learning research has been focusing more and more on interpretability \citep{gilpin2018explaining}.
However, there are many nuances to \emph{interpretability} \citep{lipton2016mythos}, and amongst them we focus on model transparency.

One strategy is to extract a simpler, interpretable model from a neural network, though this comes at the cost of performance.
For example, \citet{thrun1995extracting} extract if-then rules, while \citet{craven1996extracting} extract decision trees. 

There is also work on making word vectors more interpretable. \citet{faruqui2015sparse} make word vectors more sparse, and \citet{herbelotvecchi2015building} learn to map distributional word vectors to model-theoretic semantic vectors.

Similarly to \citet{leietal2016rationalizing}, \citet{titov2008joint} extract informative fragments of text by jointly training a classifier and a model predicting a stochastic mask, while relying on Gibbs sampling to do so. Their focus is on using the sentiment labels as a weak supervision signal for opinion summarization rather than on rationalizing classifier predictions. 

There are also related approaches that aim to interpret an already-trained model, in contrast to \citet{leietal2016rationalizing} and our approach where the rationale is jointly modeled. \citet{ribeiro2016why} make any classifier interpretable by approximating it locally with a linear proxy model in an approach called LIME, and \citet{alvarezmelis2017causalexplaining} propose a framework that returns input-output pairs that are causally related. 

\paragraph{Learning from rationales.} 
Our work is different from approaches that aim to improve classification using rationales as an additional input \citep{zaidan2007using,zaidan-eisner-2008-modeling,zhang-etal-2016-rationale}. Instead, our rationales are latent and we are interested in uncovering them. We only use annotated rationales for evaluation.

\paragraph{Sparse layers.} 
Also arguing for enhanced interpretability, \citet{niculae2017regularized} propose a framework for learning sparsely activated attention layers based on smoothing the $\max$ operator. They derive a number of relaxations to $\max$, including $\softmax$ itself, but in particular, they target relaxations such as $\sparsemax$ \citep{martins2016softmax} which, unlike $\softmax$, are sparse (i.e. produce vectors of probability values with components that evaluate to exactly $0$). 
Their activation functions are themselves solutions to convex optimization problems, to which they provide efficient forward and backward passes.
The technique can be seen as a deterministic sparsely activated layer which they use as a drop-in replacement to standard attention mechanisms.
In contrast, in this paper we focus on binary outcomes rather than $K$-valued ones. 
\citet{niculaeetal2018sparselatent} extend the framework to structured discrete spaces where they learn sparse parameterizations of discrete latent models. In this context, parameter estimation requires exact marginalization of discrete variables or gradient estimation via REINFORCE. 
They show that oftentimes distributions are sparse enough to enable exact marginal inference.

\citet{pengthomsonsmith2018spigot} propose SPIGOT, a proxy gradient to the non-differentiable $\argmax$ operator. This proxy requires an $\argmax$ solver (e.g. Viterbi for structured prediction) and, like the straight-through estimator \citep{bengio2013estimating}, is a biased estimator. Though, unlike ST it is efficient for structured variables. In contrast, in this work we chose to focus on unbiased estimators.

\paragraph{Rectified Distributions.} The idea of rectified distributions has been around for some time. The rectified Gaussian distribution \citep{RectGaussian}, in particular, has found applications to factor analysis \citep{harva2005variational} and approximate inference in graphical models \citep{winn2005variational}. 
\citet{louizos2017learning} propose to stretch and rectify samples from the BinaryConcrete (or GumbelSoftmax) distribution \citep{MaddisonEtAl2017:Concrete,JangEtAl2017:GumbelSoftmax}.
They use rectified variables to induce sparsity in parameter space via a relaxation to $L_0$. We adapt their technique to promote sparse \emph{activations} instead. 
\citet{rolfe2016discrete} learns a relaxation of a discrete random variable based on a tractable mixture of a point mass at zero and a continuous reparameterizable density, thus enabling reparameterized sampling from the half-closed interval $[0, \infty)$. In contrast, with HardKuma we focused on giving support to both 0s and 1s.
\section{Conclusions}
We  presented a differentiable approach to extractive rationales, including an objective that allows for specifying how much text is to be extracted.
To allow for reparameterized gradient estimates and support for binary outcomes we introduced the HardKuma distribution. 
Apart from extracting rationales, we showed that HardKuma has further potential uses, which we demonstrated on premise-hypothesis attention in SNLI.
We leave further explorations for future work.

\section*{Acknowledgments}
We thank Luca Falorsi for pointing us to \citet{louizos2017learning}, which inspired the HardKumaraswamy distribution.
This work has received funding from the European Research Council (ERC StG BroadSem 678254), the European Union's Horizon 2020 research and innovation programme (grant agreement No 825299, GoURMET), and the Dutch National Science Foundation (NWO VIDI 639.022.518, NWO VICI 277-89-002).



\bibliography{ms}
\bibliographystyle{acl_natbib}

\clearpage
\appendix

\section{Kumaraswamy distribution}
\label{app:kuma}

\begin{figure}[tbh]
    \centering
    \includegraphics[width=\columnwidth]{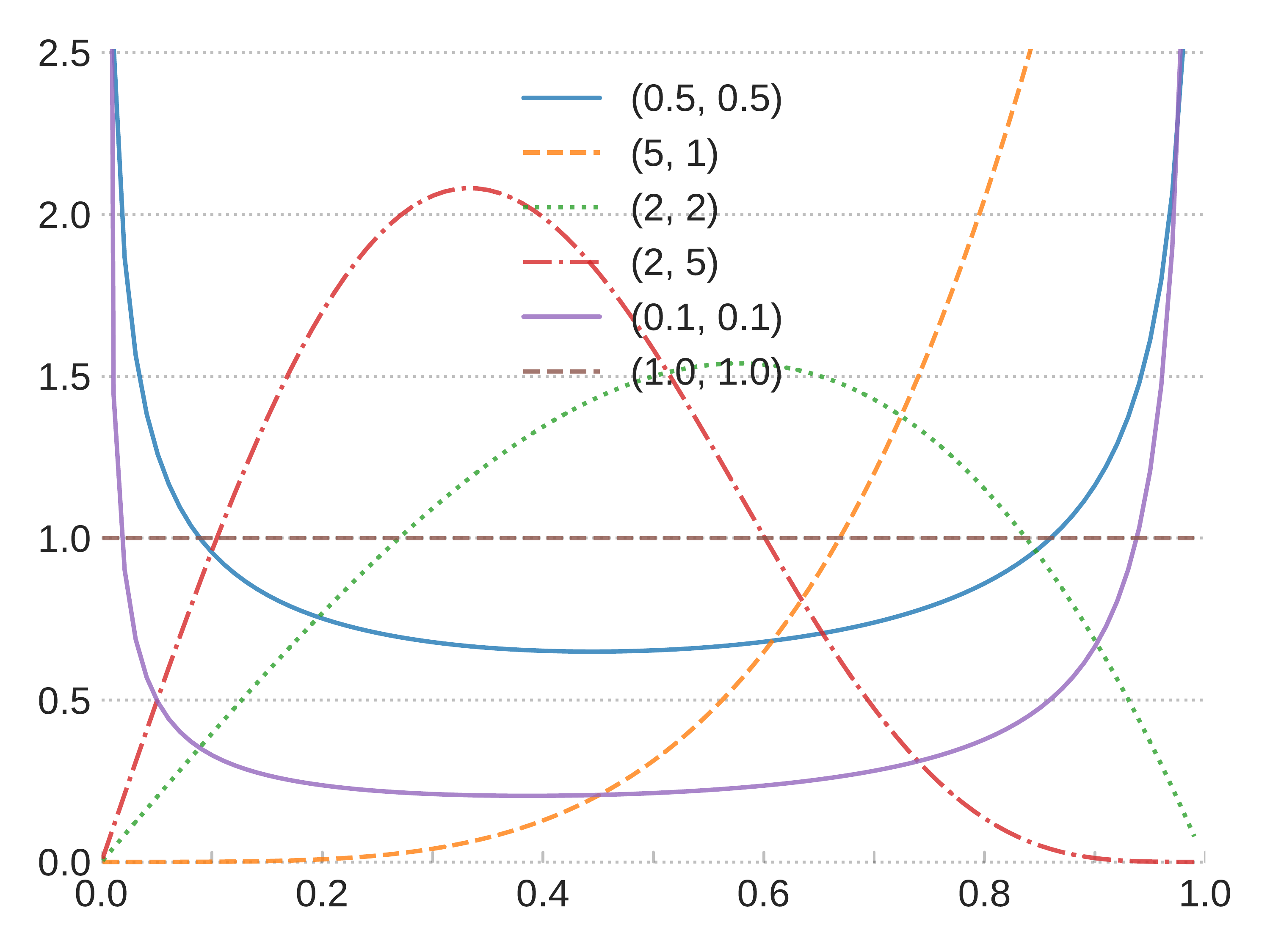}
    \caption{Kuma plots for various (a, b) parameters.}
    \label{fig:kumas}
\end{figure}

A Kumaraswamy-distributed variable $\Krv \sim \Kuma(a, b)$ takes on values in the open interval $(0, 1)$ and has density
\begin{equation}
    f_\Krv(\krv;a, b) = a b \krv^{a -1 }(1 - \krv^{a})^{b-1} ~,
\end{equation}
where $a \in \mathbb R_{>0}$ and $b \in \mathbb R_{>0}$ are shape parameters.
Its cumulative distribution takes a simple closed-form expression
\begin{subequations}\label{eq:kumacdf}
\begin{align}
    F_\Krv(\krv; a, b) &= \int_0^\krv f_\Krv(\xi|a, b) \dd \xi  \\
    &= 1-(1-\krv^{a})^{b} ~,
\end{align}
\end{subequations}
with inverse 
\begin{equation}\label{eq:invcdf}
    F_\Krv^{-1}(u; a, b) = \left(1- (1-u)^{\sfrac{1}{b}}\right)^{\sfrac{1}{a}}  ~.
\end{equation}

\subsection{\label{app:stretched}Generalised-support Kumaraswamy}

We can generalise the support of a Kumaraswamy variable by specifying two constants $l < r$ and transforming a random variable $\Krv \sim \Kuma(a, b)$ to obtain $\Trv \sim \Kuma(a, b, l, r)$  as shown in (\ref{eq:stretch}, left).
\begin{align}\label{eq:stretch}
    \trv = l + (r - l) \krv & & \krv = \sfrac{(\trv - l)}{(r - l)}
\end{align}
The density of the resulting variable is 
\begin{subequations}
\begin{align}
    &f_\Trv(\trv; a, b, l, r) \\
    &= f_\Krv\left(\textstyle\frac{t-l}{r-l}; a, b\right) \abs{\dv{\krv}{\trv}} \\
    &= f_\Krv\left(\textstyle\frac{t-l}{r-l}; a, b\right) \frac{1}{(r - l)}
\end{align}
\end{subequations}
where $r - l > 0$ by definition.
This \emph{affine} transformation leaves the cdf unchanged, i.e. 
\begin{equation}\label{eq:app:trv_cdf}
\begin{aligned}
    &F_\Trv(\trv_0; a, b, l, r) = \int_{-\infty}^{\trv_0} f_\Trv(\trv; a, b, l, r) \dd \trv \\
    &= \int_{-\infty}^{\trv_0} f_K\left(\textstyle\frac{\trv-l}{r-l}; a, b\right)\frac{1}{(r-l)} \dd \trv \\
    &= \int_{-\infty}^{\frac{\trv_0 - l}{r - l}} f_\Krv(\krv; a, b)\frac{1}{(r-l)} (r-l) \dd \krv \\
    &= F_\Krv\left(\textstyle\frac{\trv_0-l}{r-l}; a, b\right)~.
\end{aligned}
\end{equation}
Thus we can obtain samples from this generalised-support Kumaraswamy by sampling from a uniform distribution $\mathcal U(0, 1)$, applying the inverse transform (\ref{eq:invcdf}),  then shifting and scaling the sample according to (\ref{eq:stretch}, left).

\subsection{\label{app:HKuma}Rectified Kumaraswamy}

First, we stretch a Kumaraswamy distribution to include $0$ and $1$ in its support, that is, with $l < 0$ and $r > 1$, we define $\Trv \sim \Kuma(a, b, l, r)$. Then we apply a \emph{hard-sigmoid} transformation to this variable, that is, $\hrv = \min(0, \max(1, \trv))$, which results in a \emph{rectified} distribution which gives support to the closed interval $[0, 1]$.
We denote this rectified variable by 
\begin{equation}
    \Hrv \sim \HKuma(a, b, l, r)
\end{equation}
whose distribution function is
\begin{equation}
\begin{aligned}
    &f_\Hrv(\hrv; a, b, l, r) = \\
    &~\mathbb P(\hrv = 0) \delta(\hrv) + \mathbb P(\hrv = 1)  \delta(\hrv - 1) \\
    &+ \mathbb P(0 < \hrv < 1) \frac{f_\Trv(\hrv; a, b, l, r) \mathds 1_{(0,1)}(\hrv)}{\mathbb P(0 < \hrv < 1)}
\end{aligned}
\end{equation}
where 
\begin{equation}
\begin{aligned}
    &\mathbb P(\hrv = 0) = \mathbb P(\trv \le 0) \\
    &= F_\Trv(0; a, b, l, r) = F_\Krv(\sfrac{-l}{(r-l)}; a, b) 
\end{aligned}
\end{equation}
is the probability of sampling exactly $0$, where
\begin{equation}
\begin{aligned}
    &\mathbb P(\hrv = 1) = \mathbb P(\trv \ge 1) = 1 - \mathbb P(\trv < 1) \\
    &= 1 - F_\Trv(1; a, b, l, r) \\
    &= 1 - F_\Krv(\sfrac{(1-l)}{(r-l)}; a, b) 
\end{aligned}
\end{equation}
is the probability of sampling exactly 1, and
\begin{equation}
 \mathbb P(0 < \hrv < 1) = 1 - \mathbb P(\hrv = 0) - \mathbb P(\hrv = 1) 
\end{equation}
is the probability of drawing a continuous value in $(0, 1)$.
Note that we used the result in (\ref{eq:app:trv_cdf}) to express these probabilities in terms of the tractable cdf of the original Kumaraswamy variable.

\subsection{\label{app:grad}Reparameterized gradients}

Let us consider the case where we need derivatives of a function $\mathcal L(u)$ of the underlying uniform variable $u$, as when we compute reparameterized gradients in variational inference. 
Note that 
\begin{equation}
    \pdv{\mathcal L}{u} = \pdv{\mathcal L}{\hrv} \times \pdv{\hrv}{\trv} \times \pdv{\trv}{\krv} \times \pdv{\krv}{u} ~, 
\end{equation}
by chain rule. 
The term $\pdv{\mathcal L}{\hrv}$ depends on a differentiable observation model and poses no challenge; the term $\pdv{\hrv}{\trv}$ is the derivative of the hard-sigmoid function, which is $0$ for $\trv < 0$ or $\trv > 1$, $1$ for $0 < \trv < 1$, and undefined for $\trv \in \{0, 1\}$; the term $\pdv{\trv}{\krv} = r - l$ follows directly from (\ref{eq:stretch}, left); the term $\pdv{\krv}{u} = \pdv{u} F^{-1}_\Krv(u; a, b)$ depends on the Kumaraswamy inverse cdf (\ref{eq:invcdf}) and also poses no challenge. 
Thus the only two discontinuities happen for $\trv \in \{0, 1\}$, which is a $0$ measure set under the stretched Kumaraswamy: we say this reparameterisation is differentiable \emph{almost everywhere}, a useful property which essentially circumvents the discontinuity points of the rectifier.

\subsection{HardKumaraswamy PDF and CDF}

\begin{figure}[tbh]
    \centering
    \includegraphics[width=\columnwidth]{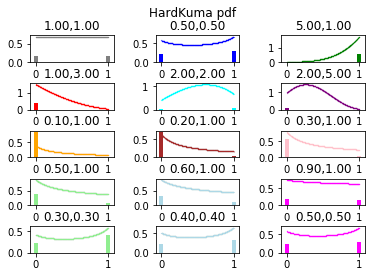}
    \caption{HardKuma pdf for various (a, b).}
    \label{fig:hardkumapdf}
\end{figure}

\begin{figure}[tbh]
    \centering
    \includegraphics[width=\columnwidth]{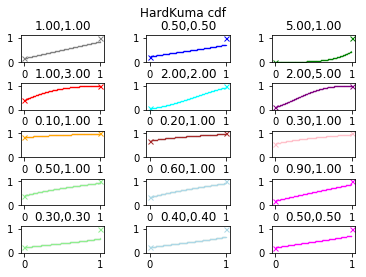}
    \caption{HardKuma cdf for various (a, b).}
    \label{fig:hardkumacdf}
\end{figure}

Figure~\ref{fig:hardkumapdf} plots the pdf of the HardKumaraswamy for various a and b parameters.
Figure~\ref{fig:hardkumacdf} does the same but with the cdf.


\section{Implementation Details}
\label{app:implementationdetails}

\subsection{Multi-aspect Sentiment Analysis}

Our hyperparameters are taken from \citet{leietal2016rationalizing} and listed in Table~\ref{tab:beer-hyperparameters}. 
The pre-trained word embeddings and data sets are available online at \url{http://people.csail.mit.edu/taolei/beer/}.
We train for 100 epochs and select the best models based on validation loss.
For the MSE trade-off experiments on all aspects combined, we train for a maximum of 50 epochs.

\begin{table}[h]
    \centering
    \begin{tabular}{lr}
    \toprule
    Optimizer     & Adam   \\
    Learning rate & 0.0004 \\ 
    Word embeddings & 200D (Wiki, fixed) \\
    Hidden size & 200 \\
    Batch size  & 256  \\
    Dropout     & 0.1, 0.2 \\
    Weight decay & $1*10^{-6}$\\
    Cell & RCNN \\
    \bottomrule
    \end{tabular}
    \caption{Beer hyperparameters.}
    \label{tab:beer-hyperparameters}
\end{table}

For the Bernoulli baselines we vary $L_0$ weight $\lambda_1$ among $\{0.0002, 0.0003, 0.0004\}$, just as in the original paper.
We set the fused lasso (coherence) weight $\lambda_2$ to $2*\lambda_1$.

For the HardKuma models we set a target selection rate to the values targeted in Table~\ref{tab:beer-results}, and optimize to this end using the Lagrange multiplier. We chose the fused lasso weight from $\{0.0001, 0.0002, 0.0003, 0.0004\}$.

\subsubsection{Recurrent Unit}

In our multi-aspect sentiment analysis experiments we use the RCNN of \citet{leietal2016rationalizing}.
Intuitively, the RCNN is supposed to capture n-gram features that are not necessarily consecutive.
We use the bigram version (filter width $n = 2$) used in \citet{leietal2016rationalizing}, which is defined as:
\begin{align*}
\vect{\lambda}_t &= \sigma(W^\lambda \vect{x}_t + U^\lambda \vect{h}_{t-1} + \vect{b}^\lambda  ) \\
\vect{c}_t^{(1)} &= \vect{\lambda}_t \odot \vect{c}_{t-1}^{(1)} + (1-\vect{\lambda}_t) \odot W_1 \vect{x}_{t} \\
\vect{c}_t^{(2)} &= \vect{\lambda}_t \odot \vect{c}_{t-1}^{(2)} + (1-\vect{\lambda}_t) \odot (\vect{c}_{t-1}^{(1)} + W_2 \vect{x}_{t} ) \\
\vect{h}_t &= \tanh (\vect{c}_t^{(2)} + \vect{b} )
\end{align*}

\subsubsection{\label{app:depz}Expected values for dependent latent variables}

The expected $L_0$ is a chain of nested expectations, and we solve each term 
\begin{equation}
\begin{aligned}
&\mathbb E_{p(z_i|x, z_{<i})}\left[ \mathbb I[z_i \neq 0] \mid z_{<i}\right] \\
&= 1 - F_K\left(\textstyle\frac{-l}{r-l}; a_i, b_i\right) 
\end{aligned}
\end{equation}
as a function of a sampled prefix, and the shape parameters $a_i, b_i = g_i(x, z_{<i}; \phi)$ are predicted in sequence.

\subsection{Sentiment Classification (SST)}

For sentiment classification we make use of the PyTorch bidirectional LSTM module for encoding sentences, for both the rationale extractor and the classifier. The BiLSTM final states are concatenated, after which a linear layer followed by a softmax produces the prediction.
Hyperparameters are listed in Table~\ref{tab:sst-hyperparameters}. 
We apply dropout to the embeddings and to the input of the output layer.

\begin{table}[h]
    \centering
    \begin{tabular}{lr}
    \toprule
    Optimizer     & Adam   \\
    Learning rate & 0.0002 \\ 
    Word embeddings & 300D Glove (fixed) \\
    Hidden size & 150 \\
    Batch size  & 25  \\
    Dropout     & 0.5 \\
    Weight decay & $1*10^{-6}$\\
    Cell & LSTM \\
    \bottomrule
    \end{tabular}
    \caption{SST hyperparameters.}
    \label{tab:sst-hyperparameters}
\end{table}

\subsection{Natural Language Inference (SNLI)}

\begin{figure*}[p]
    \centering
  \begin{subfigure}[b]{0.45\textwidth}
        \includegraphics[width=\textwidth,clip,trim=0 3cm 0 0]{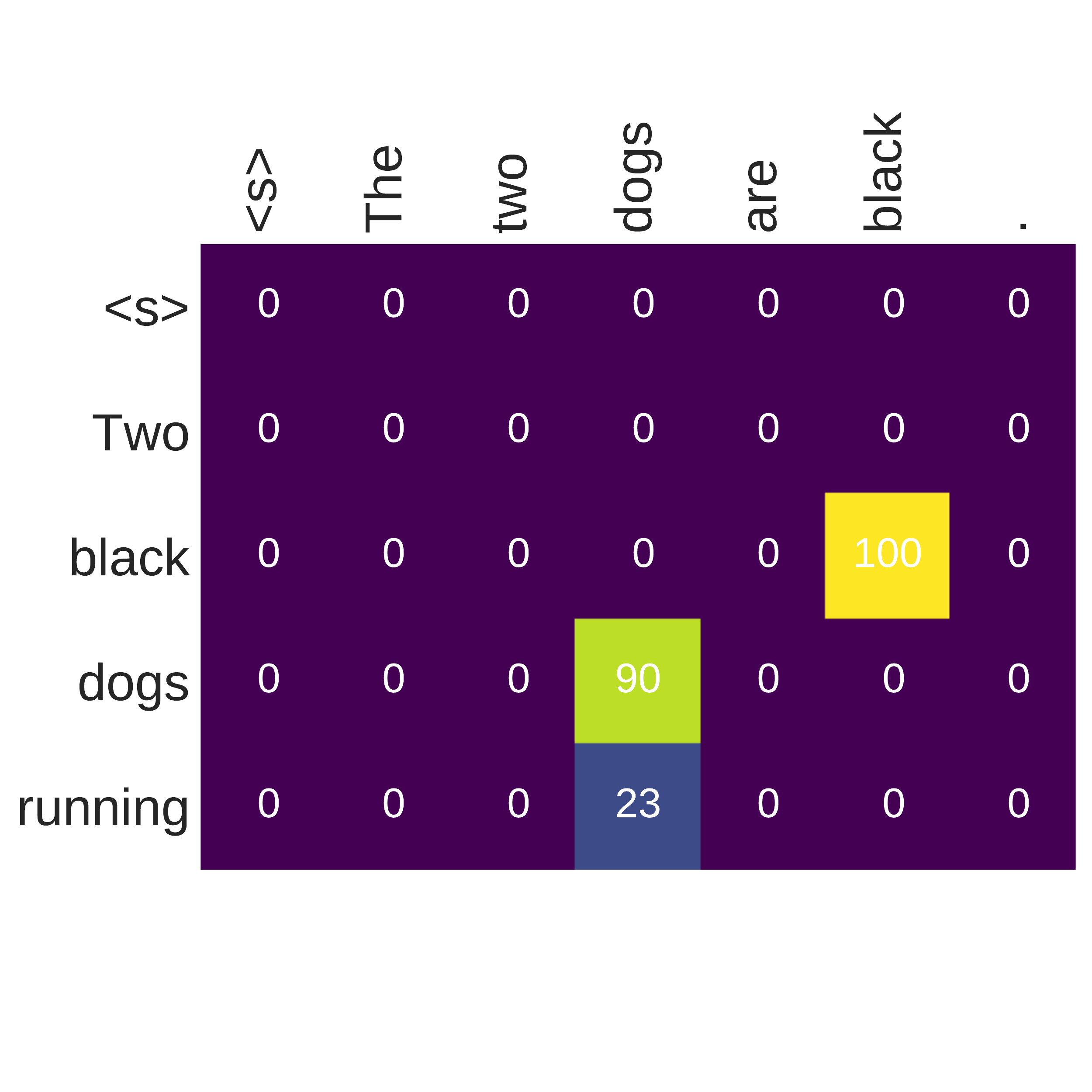}
        \caption{Entailment (correct)}
        \label{fig:hk-snli-entailment-correct}
    \end{subfigure}%
    \begin{subfigure}[b]{0.45\textwidth}
        \includegraphics[width=\textwidth,clip,trim=0 3cm 0 0]{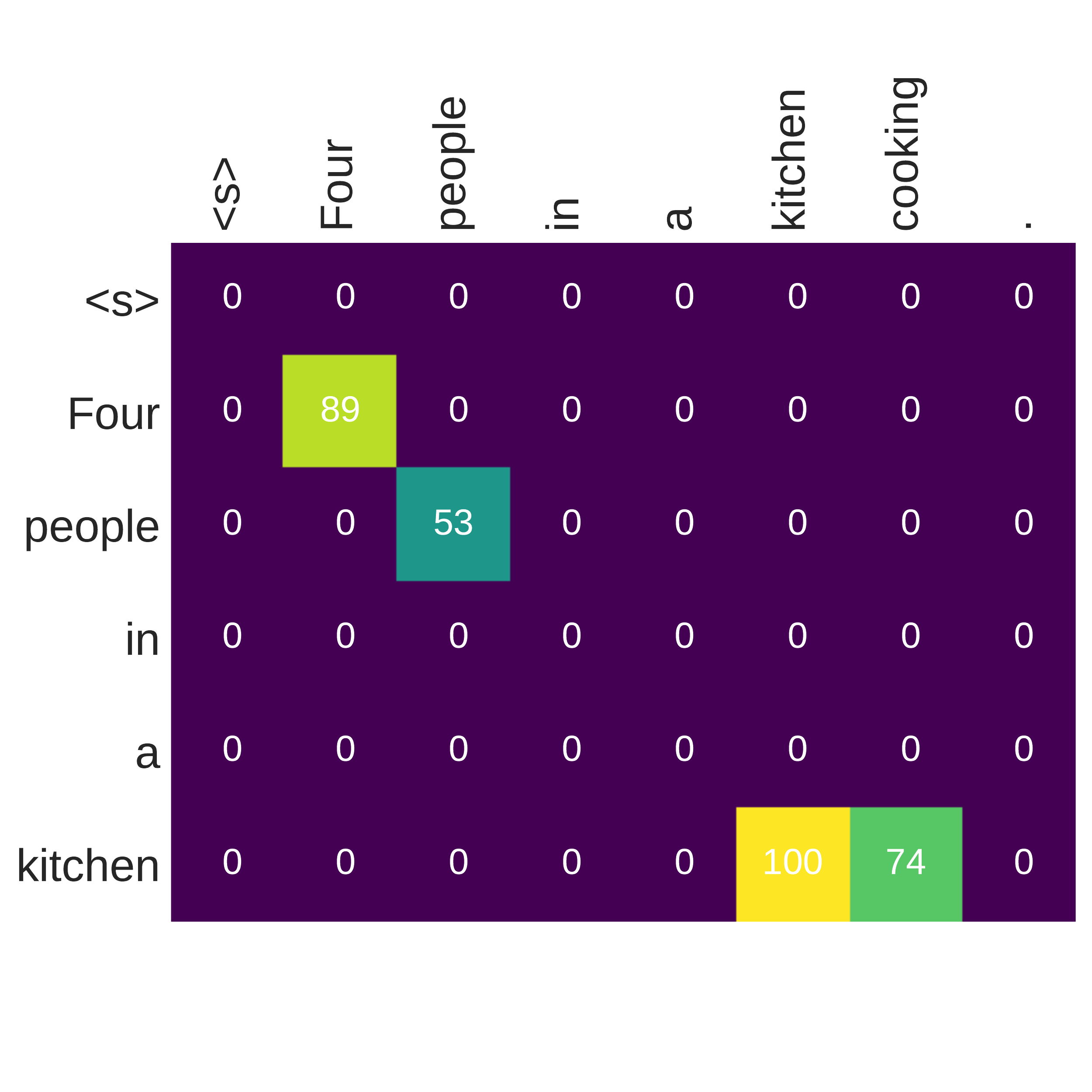}
        \caption{Entailment (incorrect, pred: neutral)}
        \label{fig:hk-snli-entailment-incorrect}
    \end{subfigure}
    \begin{subfigure}[b]{0.45\textwidth}
        \includegraphics[width=\textwidth,clip,trim=0 1cm 0 0]{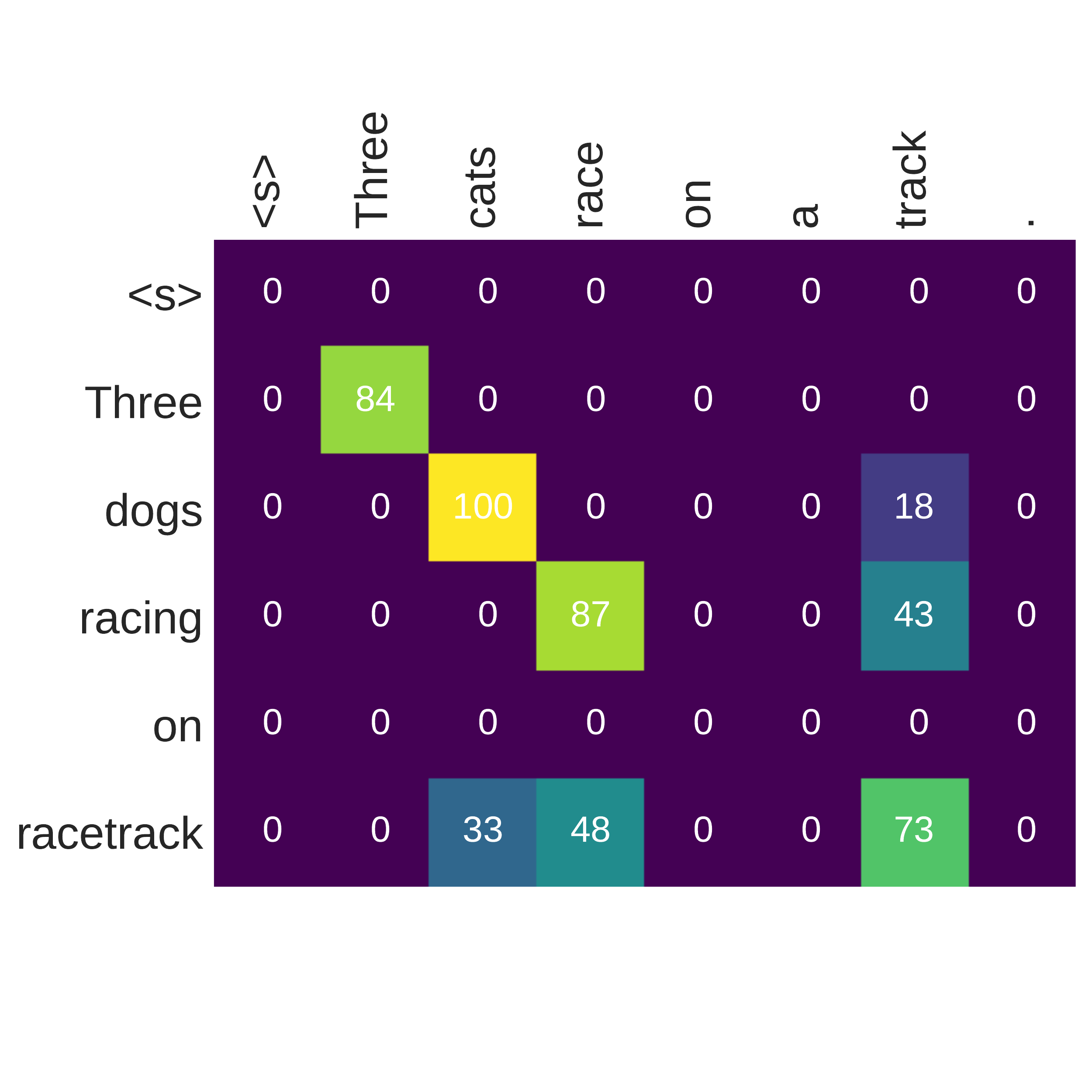}
        \caption{Contradiction (correct)}
        \label{fig:hk-snli-contradiction-correct}
    \end{subfigure}%
    \begin{subfigure}[b]{0.45\textwidth}
        \includegraphics[width=\textwidth]{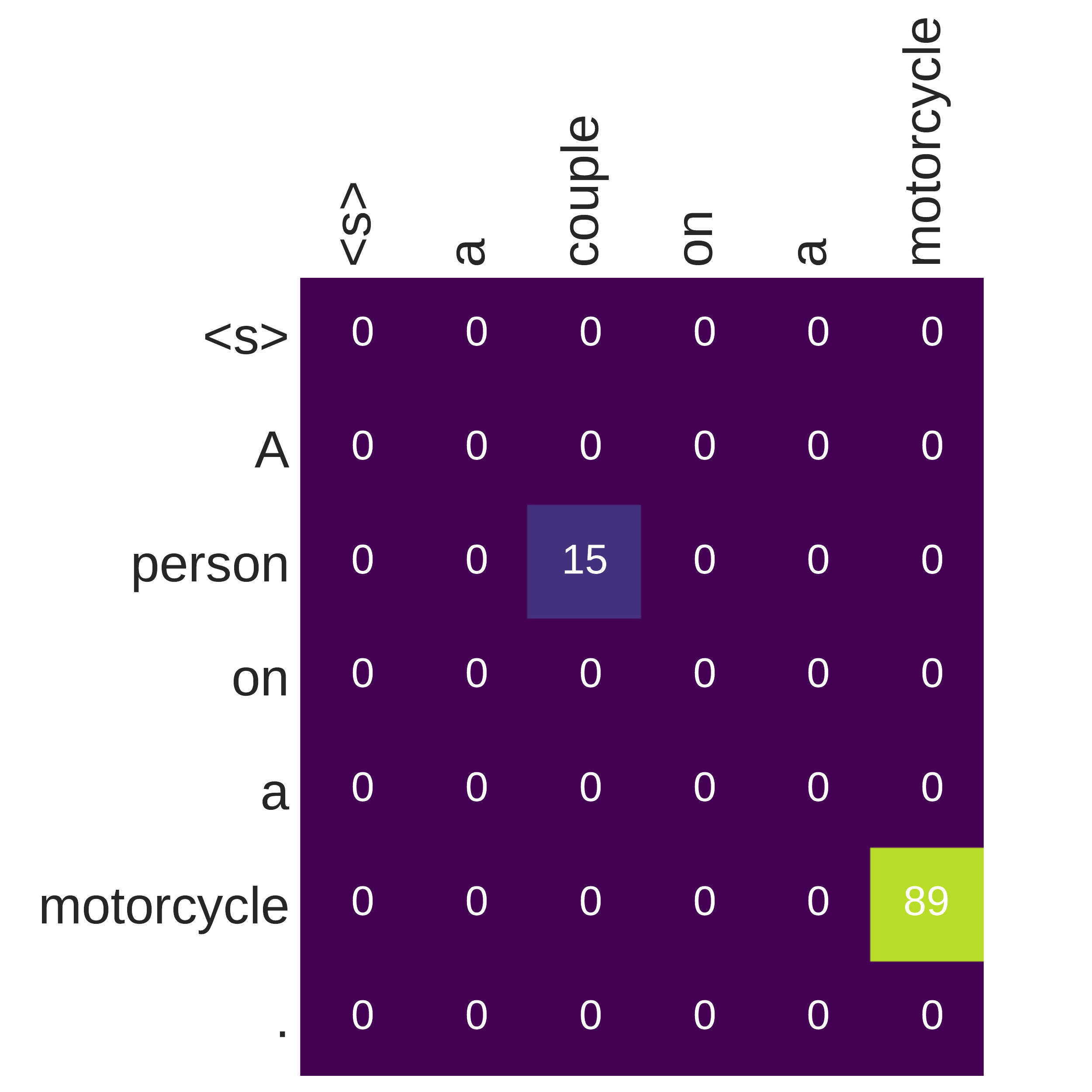}
        \caption{Contradiction (incorrect, pred: entailment)}
        \label{fig:hk-snli-contradiction-incorrect}
    \end{subfigure}
    
    \begin{subfigure}[b]{0.45\textwidth}
        \includegraphics[width=\textwidth]{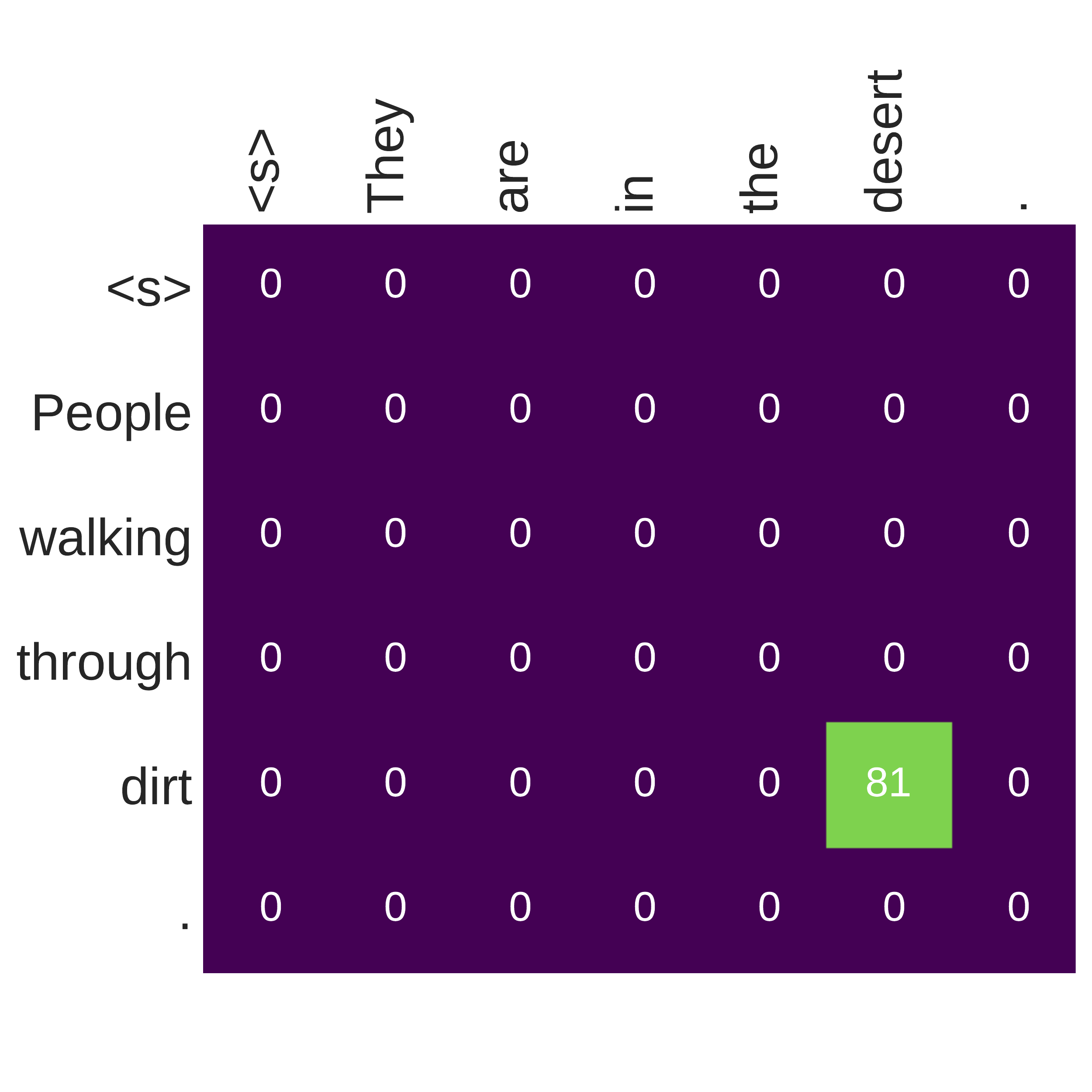}
        \caption{Neutral (correct)}
        \label{fig:hk-snli-neutral-correct}
    \end{subfigure}%
    \begin{subfigure}[b]{0.45\textwidth}
        \includegraphics[width=\textwidth]{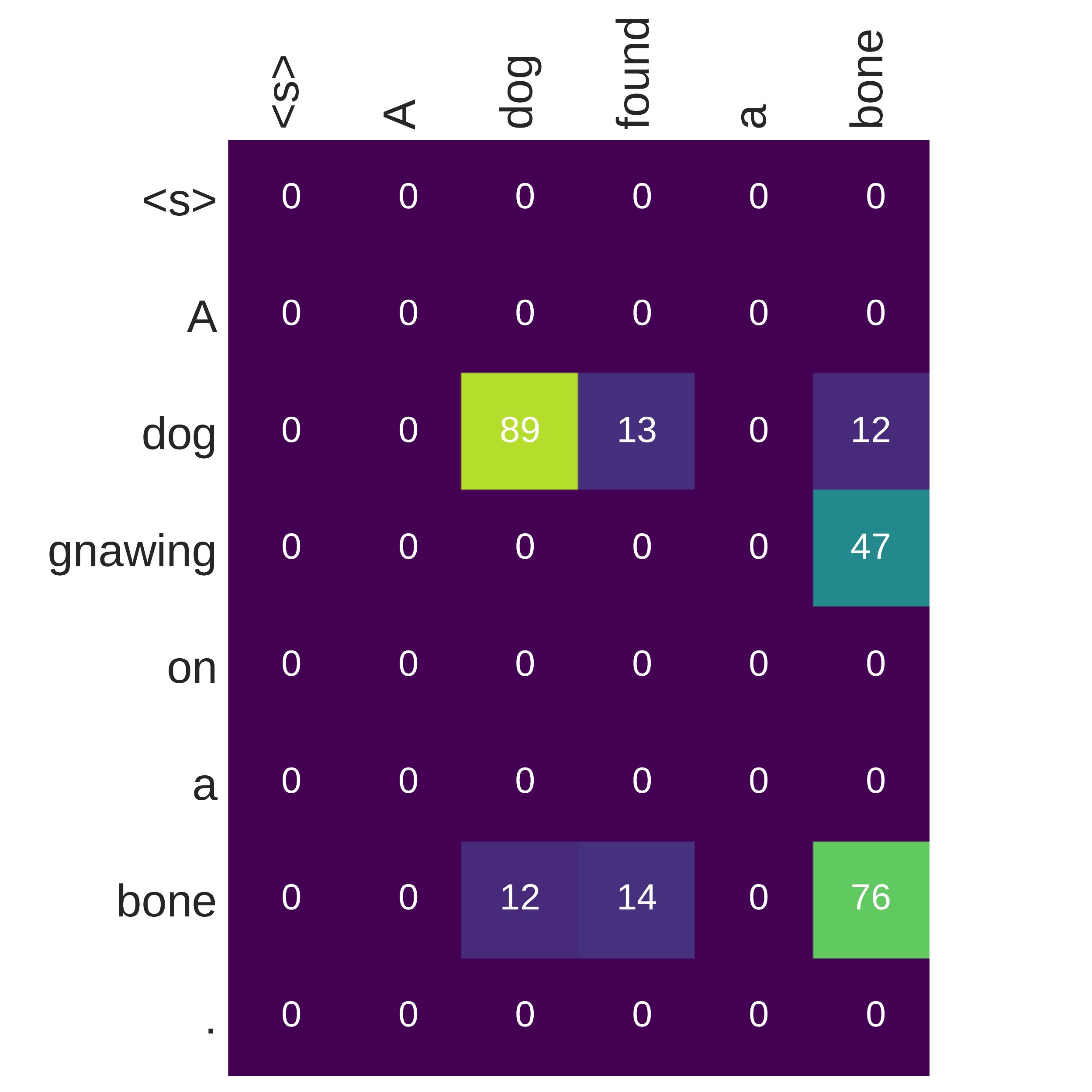}
        \caption{Neutral (incorrect, pred: entailment)}
        \label{fig:hk-snli-neutral-incorrect}
    \end{subfigure}    
    \caption{HardKuma attention in SNLI for entailment, contradiction, and neutral.}
    \label{fig:hardkuma-attention-examples}
\end{figure*}

Our hyperparameters are taken from \citet{parikhetal2016decomposable} and listed in Table~\ref{tab:snli-hyperparameters}. 
Different from \citeauthor{parikhetal2016decomposable} is that we use Adam as the optimizer and a batch size of 64.
Word embeddings are projected to 200 dimensions with a trained linear layer.
Unknown words are mapped to 100 unknown word classes based on the MD5 hash function, just as in \citet{parikhetal2016decomposable},
and unknown word vectors are randomly initialized.
We train for 100 epochs, evaluate every 1000 updates, and select the best model based on validation loss.
Figure~\ref{fig:hardkuma-attention-examples} shows a correct and incorrect example with HardKuma attention for each relation type (entailment, contradiction, neutral).

\begin{table}[h]
    \centering
    \begin{tabular}{lr}
    \toprule
    Optimizer     & Adam   \\
    Learning rate & 0.0001 \\ 
    Word embeddings & 300D (Glove, fixed) \\
    Hidden size & 200 \\
    Batch size  & 64  \\
    Dropout     & 0.2 \\
    \bottomrule
    \end{tabular}
    \caption{SNLI hyperparameters.}
    \label{tab:snli-hyperparameters}
\end{table}

\end{document}